\documentclass[runningheads]{llncs}

\usepackage[year=2026,ID=11859]{eccv}

\usepackage{eccvabbrv}

\usepackage{graphicx}
\usepackage{booktabs}
\usepackage[table,xcdraw]{xcolor}
\usepackage{multirow}
\usepackage{makecell}
\usepackage[accsupp]{axessibility}  

\usepackage{hyperref}

\usepackage{orcidlink}

\begin{document}

\title{From Sparsity to Simplicity: Enabling Simpler Sequential Replacements via Sparse Attention Distillation} 

\titlerunning{From Sparsity to Simplicity (S2S)}

\author{Yuxin Ren\inst{1}\orcidlink{0009-0005-6163-6955} \and
Maxwell D Collins\inst{2}\orcidlink{0009-0005-5759-9821} \and
Miao Hu\inst{2}\orcidlink{0000-0002-6370-5564} \and
Huanrui Yang\inst{1}\orcidlink{0000-0002-3384-4512}}

\authorrunning{Y.~Ren et al.}

\institute{University of Arizona, Tucson AZ, USA \\
\email{\{yuxinr,huanruiyang\}@arizona.edu} \and
TetraMem, Inc., San Jose CA, USA \\
\email{\{maxwell.collins,miao.hu\}@tetramem.com}}

\maketitle

\begin{abstract}
Self-attention serves as the core foundation of large-scale transformer pretraining, but its quadratic token interaction cost makes inference expensive. Replacing attention with simpler sequential modules is appealing, yet naïve substitution is often lossy, especially at larger scales. This paper revisits attention replacement through the lens of sparsity. 
Based on the observation of diverse sparsity patterns across transformer layers, we posit that pretrained transformers decompose the complex token dependency across tokens into various sequence-to-sequence mappings of diverse complexities, where some layer functionalities can be approximated and replaced with much simpler sequential modules without loss.
We evaluate this premise using a plug-and-play layer-wise distillation framework to approximate and replace attention functionalities in pretrained vision transformer models. 
Controlled group-wise replacements under a fixed training budget reveal a clear pattern: substituting layers with sparser attention incurs substantially smaller accuracy drops than replacing denser ones. 
We further impose explicit attention sparsity on the pretrained ViT via AViT-style token retention and perform sparsity-guided distillation for sequential replacing models, where we see increasing teacher sparsity consistently reduces the student–teacher gap.
The proposed method achieves efficient attention replacement for reduced parameter size and latency through the guidance of attention sparsity. 
Code is available at \url{https://github.com/aliothren/FAR}.

  
\end{abstract}    
\section{Introduction}
\label{sec:intro}
 
Transformers have become the dominant backbone across vision, language, and multimodal learning~\cite{vaswani2023attention, devlin2019bert, dosovitskiy2021vit, radford2021clip}. 
Their success is largely attributed to the self-attention mechanism, which performs flexible token mixing through all-pair interactions, providing a powerful method for modeling long-range dependencies and aggregating context adaptively~\cite{cordonnier2020relationship}. 
However, attention comes with quadratic complexity in sequence length and heavy activation-to-activation multiplications, resulting in high latency and memory overhead during inference~\cite{tay2022efficient,dao2022flash}.

This tension has motivated extensive research on efficient attention.
Most efficiency-oriented approaches modify the attention computation itself through linearized or recurrent formulations~\cite{katharopoulos2020transformers, gu2024mamba} to reduce the cost while maintaining accuracy, yet such simplifications inevitably reduce the expressive capacity of attention. 
Recent theoretical work attributes this gap to the lossy hidden-state memory of recurrent models, which prevents the exact retrieval and combination of distant contextual information~\cite{wen2025rnns}. 
This signifies why attention is indispensable during large-scale pretraining, as it uniquely supports lossless retrieval of token relationships across arbitrary context lengths. 

Nevertheless, the same level of expressivity may not be required during inference, where attention operates on already contextualized representations produced by preceding layers, and its computation may be effectively far less complex and more structured than its dense all-pair form suggests. 
Empirically, pretrained transformers exhibit substantial redundancy in attention at inference time: attention maps are often sparse and low-rank, admitting compressed structure even without explicit regularization~\cite{he2024matterstransformers, bhojanapalli2021leveragingredundancy}. 
Crucially, this sparsity is not uniform across the network. 
Different layers show markedly different token-interaction patterns, indicating that the role of attention varies across depth and that certain blocks may rely on only a small subset of token interactions.

These observations motivate a shift in how we think about attention replacement. 
Rather than asking whether a simple module can match attention everywhere, we ask when and where attention behaves like a mapping that can be replaced with minimal loss.
This leads to the central hypothesis of this paper.

\textbf{Hypothesis.}
Pretraining decomposes complex dependencies into block-wise mappings with different effective complexity, and inference-time attention sparsity is a signature of this complexity.
We therefore expect sparse blocks to be more replaceable by simpler sequential modules, and explicit sparsification of the teacher to further improve distillation.

We refer to this sparsity guided replacement principle as \textbf{Sparsity to Simplicity (S2S)}. 
S2S couples a unified replacement scheme with sparsity aware supervision, using attention sparsity to identify replaceable blocks and to distill attention behavior into simpler sequential substitutes.

We evaluate the hypothesis in a series of controlled experiments. We start by designing a unified, plug-and-play layer-wise attention distillation and replacement scheme. The scheme replaces attention blocks with sequential modules such as Mamba~\cite{gu2024mamba} and LSTMs~\cite{hochreiter1997long} that match the original input and output interface, so the backbone structure remains unchanged and pretrained weights can be reused as much as possible. 
This scheme provides a unified baseline of the feasibility of attention replacement across layers, backbones, and substitute model, where we use the performance gap before and after replacement to reflect how well a sequential module can approximate the target attention mapping. 


Guided by this, we test the link between attention sparsity and replacement feasibility across the model depth. We introduce an AViT-style token retention~\cite{Yin_2022_CVPR} as an explicit and controllable attention sparsification mechanism, which aligns with the natural depth-varying attention sparsity trend observed in pretrained ViT models. 
We replace attention blocks at different layer groups with sequential modules under identical training conditions to isolate the approximation difficulty of the replaced mapping. 
Replacements in sparser blocks consistently incur smaller accuracy drops and lower distillation error than replacements in denser blocks, supporting the view that sparsity marks simpler mappings that are easier to approximate.
Further experiments on explicitly controlling the attention sparsity level also shows that increasing teacher attention sparsity consistently shrinks the student-teacher gap, suggesting that explicit attention sparsification exposes a desirable behavior that is easier for alternative sequential modules to learn.

Following the S2S principle, we demonstrate that replacing the most replaceable 3 late-layer attentions in pretrained DeiT with designed BiMamba blocks under matched A-ViT sparsity preserves sparse-attention accuracy while improving efficiency (up to 1.71$\times$ estimated speedup at $384^2$). The effectiveness of S2S also extends to DeiT transfer benchmarks, where the replacement models can be finetuned with no clear performance gap to their attention-based counterparts.

\section{Related Work}
\label{sec:relatedwork}

\paragraph{Sparsity, Redundancy and Efficiency in Attention.}
A body of work has revealed substantial redundancy in pretrained attention at inference time.
Bhojanapalli et al.~\cite{bhojanapalli2021leveragingredundancy} and He et al.~\cite{he2024matterstransformers} show that only a small subset of heads and token interactions contribute meaningfully, while many attention weights can be pruned with limited accuracy loss.
Dynamic token pruning methods such as DynamicViT~\cite{DynamicViT2021}, TokenLearner~\cite{TokenLearner2021}, and A-ViT~\cite{Yin_2022_CVPR} exploit this redundancy by adaptively reducing token computation.
In language models, a related line of work improves long-context decoding by making inference more selective through memory and KV-cache management~\cite{xiao2024streamingllm, zhang2023ho, cai2025pyramidkv, li2024snapkv}.
These methods expose that inference behavior is often dominated by a small set of important tokens and interactions, but they still retain attention as the underlying token-mixing operator.
In contrast, we use layer-wise sparsity as a signal of where attention mappings become easier to approximate, and we study when such sparse behavior can be distilled into simpler sequential modules that replace attention itself.

\paragraph{Efficient Architectures Beyond Self-Attention.}
The quadratic cost of self-attention has motivated attention-free or attention-reduced architectures that replace all-pair token mixing with simpler mechanisms.
Examples include MLP-Mixer~\cite{tolstikhin2021mlpmixer} and related MLP-based mixers, as well as recurrent or state-space alternatives such as RetNet~\cite{sun2023retnet} and Mamba~\cite{gu2024mamba}.
While these designs improve scalability, matching standard transformers often requires careful design and training, and direct attention removal can be lossy.
Rather than proposing a new token mixer, we study when pretrained attention is replaceable and how sparsity can be leveraged to make sequential replacements closer to lossless.

\paragraph{Distillation and Modular Replacement.}
Knowledge distillation~\cite{hinton2015distilling} is widely used to transfer representations across models and has been extended to intermediate supervision and layer-wise objectives for compact models~\cite{jiao2020tinybert, sun2020mobilebert, wang2020minilm, fan2019reducing}.
Recent work also explores cross-architecture distillation, including substituting parts of transformer models with sequential modules under teacher guidance~\cite{NEURIPS2024_mambainllama}.
Different from prior efforts that focus on specific substitutions, we provide controlled, sparsity-guided analysis connecting layer-wise attention sparsity to replacement difficulty, and we introduce S2S distillation to exploit explicit sparsification when distilling attention into simpler sequential models.
\section{Methods}
\label{sec:method}

\begin{figure}[t]
    \centering
    \includegraphics[width=\linewidth]{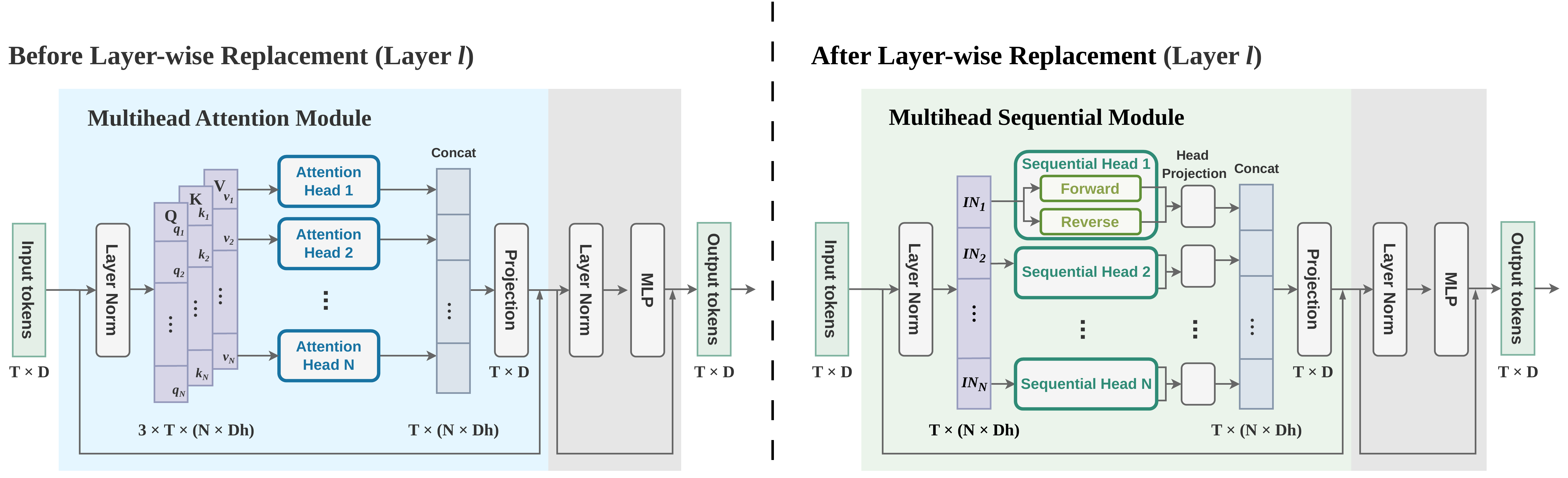}
    \caption{\textbf{Unified layer-wise attention replacement at layer $l$.}
    At a selected Transformer layer $l$, the multihead self-attention token mixer is replaced by a multihead sequential module under the same block interface and input/output dimensions, while the remaining block components (gray) are kept unchanged.}
    \label{fig:replacement_framework}
    \vspace{-5pt}
\end{figure}

\subsection{Unified Layer-wise Attention Replacement}
\label{sec:method_replacement}

Attention-based models, with Vision Transformers as a representative example, are built by stacking many identical blocks, forming a natural sequence of layer-by-layer transformations. 
This stacked structure suggests a decomposition view: large-scale pretraining can encode complex dependencies into these layer-wise mappings, so a replacement module only needs to match the mapping of the specific layers it substitutes rather than distilling the entire model globally. 
At the same time, since self-attention is crucial during pretraining, the replacement should preserve as much pretrained knowledge and backbone structure as possible. 
These considerations motivate a unified plug-and-play, layer-wise replacement setting that keeps the original layer interface intact and swaps only the attention token-mixing component in selected layers, enabling layer-selective replacement and layer-aligned supervision for learning the decomposed mappings.

\paragraph{Interface-aligned layer replacement.}
Consider a pretrained Transformer backbone with $L$ layers. 
At layer $l$, the input tokens are denoted by $\mathbf{x}^{(l)} \in \mathbb{R}^{T \times D}$, and the layer follows the standard pre-norm residual structure: a token-mixing sublayer followed by an MLP. 
As summarized in Fig.~\ref{fig:replacement_framework}, our replacement keeps the layer interface unchanged and swaps only the token-mixing component.
Concretely, a standard attention layer computes
\begin{equation}
\mathbf{y}^{(l)} = \mathbf{x}^{(l)} + \mathrm{Attn}^{(l)}\!\left(\mathrm{LN}_1(\mathbf{x}^{(l)})\right), \qquad
\mathbf{x}^{(l+1)} = \mathbf{y}^{(l)} + \mathrm{MLP}^{(l)}\!\left(\mathrm{LN}_2(\mathbf{y}^{(l)})\right),
\end{equation}
where $\mathrm{Attn}^{(l)}(\cdot)$ performs attention-based token mixing.
Our interface-aligned replacement substitutes $\mathrm{Attn}^{(l)}$ with a sequential token mixer $\mathrm{Seq}^{(l)}$ that matches the same input and output shape, yielding
\begin{equation}
\mathbf{y}^{(l)} = \mathbf{x}^{(l)} + \mathrm{Seq}^{(l)}\!\left(\mathrm{LN}_1(\mathbf{x}^{(l)})\right), \qquad
\mathbf{x}^{(l+1)} = \mathbf{y}^{(l)} + \mathrm{MLP}^{(l)}\!\left(\mathrm{LN}_2(\mathbf{y}^{(l)})\right).
\end{equation}
LayerNorm, residual connections, the MLP sublayer, and the classification head are preserved, while only the token-mixing operator is changed.
This design maximizes reuse of pretrained parameters outside the replaced token mixer, and enables layer-selective replacement: any subset of layers can be swapped under the same backbone, allowing controlled comparisons across depths and targeted replacement when only certain layers are intended to change.

\paragraph{Multihead bidirectional sequential token mixer.}
Self-attention mixes tokens through multiple heads, enabling different subspaces to capture different interaction patterns. 
To preserve this head-wise inductive bias and maintain the original layer interface, the sequential token mixer $\mathrm{Seq}^{(l)}$ adopts the same multihead wrapper. 
Given the normalized input $\mathbf{u}=\mathrm{LN}_1(\mathbf{x}^{(l)})\in\mathbb{R}^{T\times D}$, a linear projection splits features into $N$ head subspaces with $D=N\cdot D_h$:
\begin{equation}
\mathbf{I}=\mathbf{u}\mathbf{W}_{\mathrm{in}}^{(l)}\in\mathbb{R}^{T\times D},\qquad
\mathbf{I}=[\mathbf{I}_1;\ldots;\mathbf{I}_N],\ \mathbf{I}_n\in\mathbb{R}^{T\times D_h}.
\end{equation}
Each head applies a \emph{bidirectional} sequential operator $\mathcal{R}$, since ViT attention is non-causal and mixes tokens using full context at every layer:
\begin{equation}
\mathbf{U}_n^{\mathrm{fwd}}=\mathcal{R}(\mathbf{I}_n),\qquad
\mathbf{U}_n^{\mathrm{rev}}=\mathcal{R}(\mathrm{Reverse}(\mathbf{I}_n)),
\end{equation}
followed by concatenation and a per-head projection back to $D_h$,
\begin{equation}
\mathbf{U}_n=[\mathbf{U}_n^{\mathrm{fwd}};\mathbf{U}_n^{\mathrm{rev}}]\in\mathbb{R}^{T\times 2D_h},\qquad
\tilde{\mathbf{U}}_n=\mathbf{U}_n\mathbf{P}_n^{(l)}\in\mathbb{R}^{T\times D_h}.
\end{equation}
Finally, head outputs are concatenated and projected to produce $\mathrm{Seq}^{(l)}(\mathbf{u})$:
\begin{equation}
\tilde{\mathbf{U}}=\mathrm{Concat}(\tilde{\mathbf{U}}_1,\ldots,\tilde{\mathbf{U}}_N)\in\mathbb{R}^{T\times D},\qquad
\mathrm{Seq}^{(l)}(\mathbf{u})=\tilde{\mathbf{U}}\mathbf{W}_{\mathrm{out}}^{(l)}\in\mathbb{R}^{T\times D}.
\end{equation}
This splitter--bidirectional--merger template keeps the surrounding layer wiring unchanged while enabling head-wise sequential token mixing.

\paragraph{Instantiations of $\mathcal{R}$.}
We instantiate $\mathcal{R}$ with two representative sequential modules to test whether the replacement behavior is specific to a particular module or reflects a more general property of sequential token mixers. 
\emph{BiMamba} adopts Mamba as a modern state-space operator with strong empirical performance, serving as a competitive sequential instantiation under the same bidirectional, multihead wrapper. 
\emph{BiLSTM} uses a bidirectional LSTM with hidden size $D_h$ per direction as a more traditional and structurally simpler baseline, probing whether the same replacement trends persist beyond contemporary state-space designs. 
For BiMamba, we simplify the official Mamba-2 implementation by removing the additional channel-expansion branch, since this auxiliary feature-mixing path acts as an extra enhancement beyond the core sequential update and is orthogonal to the token-mixing behavior targeted by our replacement.
With this design, both instantiations share the identical multihead wrapper and projections, differing only in the choice of the per-head operator $\mathcal{R}$.

\paragraph{Layer-wise supervision enabled by replacement.}
Throughout, the \emph{teacher} refers to the pretrained Transformer with attention, and the \emph{student} refers to the model where selected layers are replaced by sequential modules.
Layer-wise replacement provides a natural interface for layer-aligned supervision: the output of each replaced token mixer can be compared to its teacher counterpart at the same layer.
A simple choice is a mean squared error similarity objective.
Let $\mathbf{z}^{(l)}_T, \mathbf{z}^{(l)}_S \in \mathbb{R}^{T \times D}$ denote the teacher and student outputs of the token-mixing sublayer at layer $l$.
The similarity loss is defined as the sum over all replaced layers $\mathcal{R}$:
\begin{equation}
\mathcal{L}_{\mathrm{sim}} \;=\; \sum_{l \in \mathcal{R}} \frac{1}{TD}\left\| \mathbf{z}^{(l)}_S - \mathbf{z}^{(l)}_T \right\|_2^2.
\end{equation}
This layer-aligned objective can be applied to any selected replacement set $\mathcal{R}$ to provide fine-grained distillation signals for learning the decomposed mappings.

\begin{figure}[t]
    \centering
    \includegraphics[width=\linewidth]{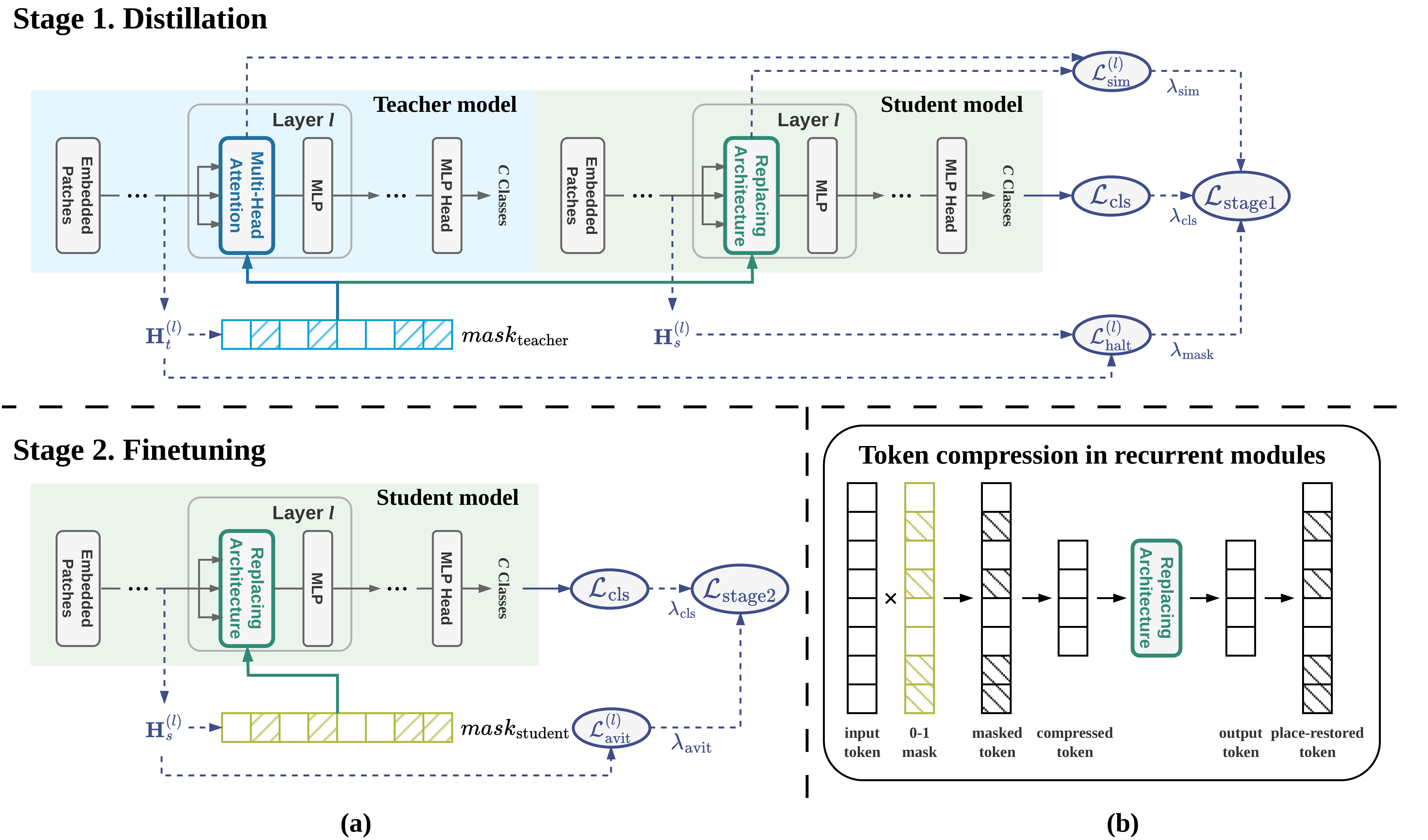}
    \caption{
        \textbf{Sparsity-guided distillation with layer-wise token masks.}
        (a) In Stage 1, the AViT teacher provides layer-wise token masks to guide teacher--student distillation; in Stage 2, the student uses its own learned masks for finetuning. 
        (b) Token compression in the sequential module, where masked tokens are temporarily removed before sequential computation and restored afterward.
    }
    \label{fig:sparse_distill}
\end{figure}

\subsection{Sparsity-Guided Distillation}
\label{sec:method_sparsity}

The replacement framework in Sec.~\ref{sec:method_replacement} allows attention layers to be substituted in a controlled, layer-wise manner. 
To further study how sparsity affects replaceability, however, naturally occurring sparsity alone is not sufficient: it is observable, but not directly controllable. 
We therefore introduce an explicit token-sparsification mechanism so that the amount and profile of sparsity can be adjusted while keeping the replacement setting unchanged. 
For this purpose, we adopt A-ViT~\cite{Yin_2022_CVPR}, which produces adaptive layer-wise token masks through halting scores. 
As shown later in Sec.~\ref{sec:exp_explicit_sparse}, the sparsity induced by A-ViT correlates well with the naturally observed layer-wise sparsity trend, making it a suitable tool for controlled sparsity studies.

More importantly, explicit sparsity must remain comparable under different token-mixing setups. 
If the teacher and student operate on different active tokens, the discrepancy no longer reflects only the difficulty of approximating the token-mixing mapping, but is entangled with token routing differences. 
We therefore couple A-ViT with the replacement framework and construct a sparsity-guided distillation scheme in which teacher and student are conditioned on aligned layer-wise token masks. 
The overall pipeline is illustrated in Fig.~\ref{fig:sparse_distill}.
Under this design, different substitutes are compared under the same sparsity pattern, so the effect of sparsity on replacement can be examined directly. 

\paragraph{A-ViT masks as controllable sparsity.}
For each layer $l$, the A-ViT teacher produces a halting score for every token and converts it into a binary retention mask.
Denote the cumulative halting probability of token $t$ at layer $l$ by $R_t^{(l)}$, with $R_t^{(0)}=0$. 
Following~\cite{Yin_2022_CVPR}, it is updated as
\begin{equation}
R_t^{(l)} = R_t^{(l-1)} + \left(1 - R_t^{(l-1)}\right) h_t^{(l)},
\end{equation}
where $h_t^{(l)} \in [0,1]$ is the halting score.
A token remains active while $R_t^{(l)} < 1$, yielding the binary mask
\begin{equation}
m_t^{(l)} = \mathbb{I}\!\left[R_t^{(l)} < 1\right].
\end{equation}
This gives a controllable layer-wise sparsity profile across the network.
When needed, the standard A-ViT regularization loss $\mathcal{L}_{\mathrm{avit}}$ is applied to stabilize the halting behavior and enforce meaningful token retention patterns.

\paragraph{Teacher-mask distillation.}
Let the teacher be a pretrained A-ViT model, and let the student be the corresponding replacement model in which selected attention layers are substituted by sequential token mixers.
In the first stage, the teacher provides the layer-wise token masks, and both models operate on the same active tokens at each replaced layer.
This shared-mask setting removes token-routing differences from the comparison and makes the supervision focus on the token-mixing function itself.
Under teacher masks, the student is trained with the layer-wise similarity loss introduced in Sec.~\ref{sec:method_replacement}, together with the task loss and a halting alignment term:
\begin{equation}
\mathcal{L}_{\mathrm{stage1}}
=
\lambda_{\mathrm{sim}} \mathcal{L}_{\mathrm{sim}}
+
\lambda_{\mathrm{mask}} \mathcal{L}_{\mathrm{halt}}
+
\lambda_{\mathrm{cls}} \mathcal{L}_{\mathrm{cls}}.
\end{equation}
Let $h^{(l)}_{t,T}$ and $h^{(l)}_{t,S}$ denote the teacher and student halting scores for token $t$ at layer $l$.
We define the halting alignment loss as
\begin{equation}
\mathcal{L}_{\mathrm{halt}}
=
\sum_{l \in \mathcal{R}} \frac{1}{T} \sum_{t=1}^{T}
\left(h^{(l)}_{t,S} - h^{(l)}_{t,T}\right)^2,
\end{equation}
where $\mathcal{R}$ is the set of replaced layers.
$\mathcal{L}_{\mathrm{cls}}$ is the standard classification task loss at the model output.
This stage directly measures how well a sequential substitute can approximate an attention layer under a prescribed sparsity pattern.

\paragraph{Student-mask finetuning.}
After the student has learned under teacher-controlled sparsity, the teacher masks are removed and the student uses its own halting modules to generate masks.
The second stage then finetunes the student with
\begin{equation}
\mathcal{L}_{\mathrm{stage2}}
=
\lambda_{\mathrm{avit}} \mathcal{L}_{\mathrm{avit}}
+
\lambda_{\mathrm{cls}} \mathcal{L}_{\mathrm{cls}},
\end{equation}
where $\mathcal{L}_{\mathrm{avit}}$ is the halting regularization loss introduced in A-ViT:
\begin{equation}
\mathcal{L}_{\mathrm{avit}}
=
\frac{1}{T}\sum_{t=1}^{T} \left|R_t^{(L)} - 1\right|
+
\lambda_{\mathrm{halt}} \sum_{l=1}^{L}
\mathrm{KL}\!\left(p^{(l)} \,\|\, p_0\right).
\end{equation}
Here $R_t^{(L)}$ is the final cumulative halting probability of token $t$, $p^{(l)}$ denotes the empirical distribution of active tokens at layer $l$, and $p_0$ is a reference prior.
This stage allows the replacement model to maintain task performance while operating under its own adaptive sparsity, and evaluates whether the sequential substitute can sustain a comparable sparsity structure when deployed autonomously.

\paragraph{Mask-guided token compression.}
Before sequential computation, inactive tokens are temporarily removed and restored afterward. 
This preserves the correct token order, avoids interference from intermediate zeros, and ensures that recurrent computation is applied only to active tokens.
\section{Experiment results}
\label{sec:experiments}

Unless otherwise specified, all experiments are conducted on ImageNet using DeiT backbones~\cite{touvron2021deit} of different scales.
The sequential replacements are instantiated with both BiMamba and BiLSTM under the same multihead wrapper described in Sec.~\ref{sec:method_replacement}.
Our evaluation focuses on three aspects: replacement quality, measured by classification accuracy; sparsity behavior, analyzed through token dependency patterns and A-ViT masks; and practical efficiency, measured by wall-clock throughput.
Implementation details and complete training hyperparameters are provided in the supplementary materials.

\begin{figure}[t]
\centering
\includegraphics[width=0.9\linewidth]{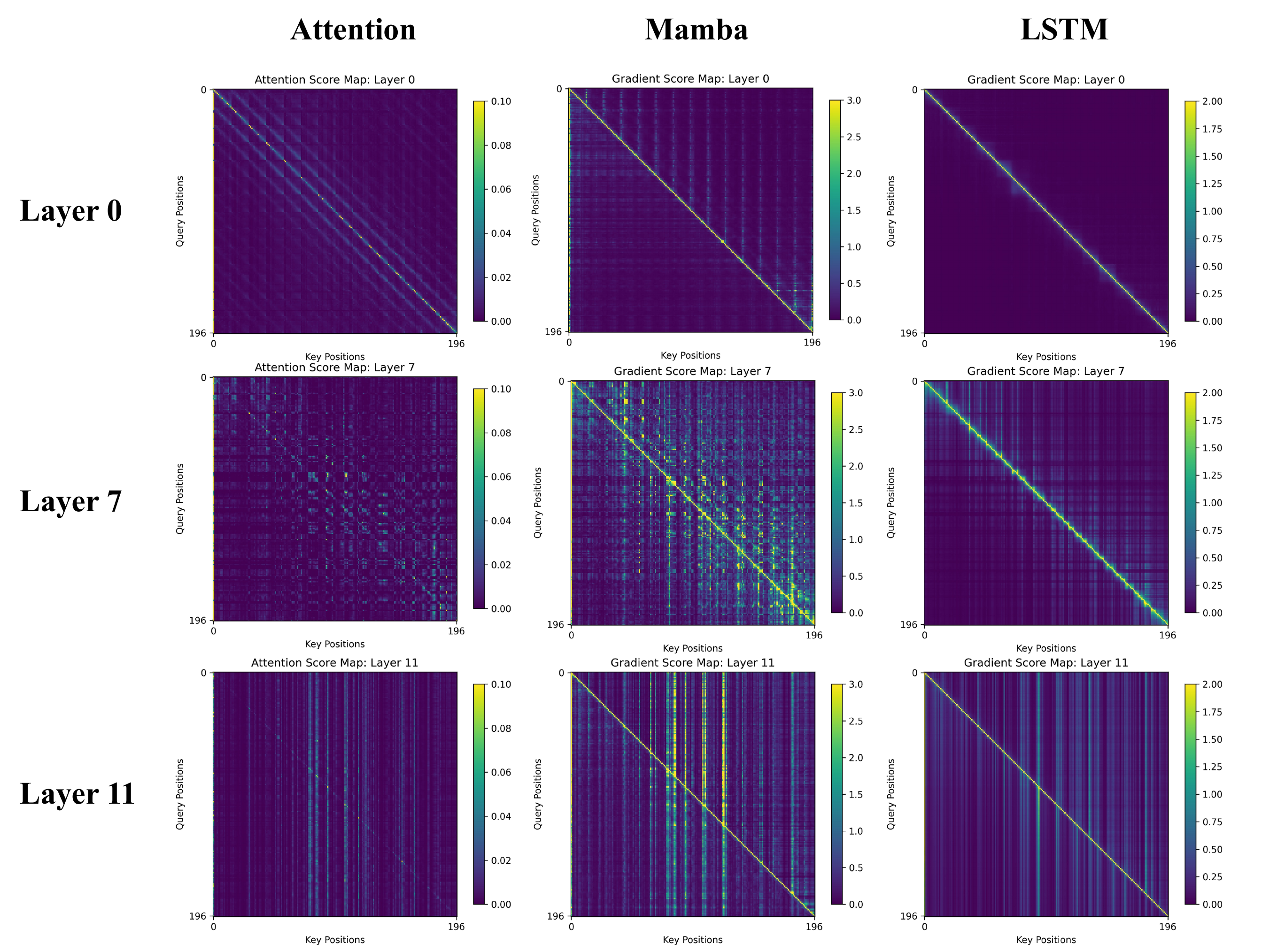}
\caption{\textbf{Cross-layer and cross-architecture token-interaction maps.}
Visualization of representative layers across Attention-, Mamba- and LSTM-based models, with Tiny-scale models shown as an example. 
DeiT is visualized by attention scores, while the sequential replacements are visualized by gradient-based interaction maps.}
\label{fig:natural_sparse}
\end{figure}

\begin{table}[t]
\centering
{
\setlength{\tabcolsep}{7pt}
\caption{\textbf{Layer-group replacement on ImageNet.}
Top-1 accuracy (\%) when replacing different layer groups under identical training conditions.
\textit{None} denotes the unmodified DeiT baseline, and \textit{Full} denotes replacing all attention layers.}
\label{tab:layer_group_replace}

\begin{tabular}{lcccccc}
\toprule
& \multicolumn{2}{c}{S2S-Tiny} & \multicolumn{2}{c}{S2S-Small} & \multicolumn{2}{c}{S2S-Base} \\
\cmidrule(lr){2-3} \cmidrule(lr){4-5} \cmidrule(lr){6-7}
Replaced layers & Mamba & LSTM & Mamba & LSTM & Mamba & LSTM \\
\midrule
None  & \multicolumn{2}{c}{72.2} & \multicolumn{2}{c}{79.8} & \multicolumn{2}{c}{81.8} \\
0--2  & 71.5 & 66.9 & 79.4 & 72.3 & 81.4 & 73.4 \\
3--5  & 70.8 & 69.0 & 79.3 & 72.6 & 81.0 & 74.1 \\
6--8  & 71.3 & 70.1 & 78.8 & 73.1 & 81.1 & 74.9 \\
9--11 & \textbf{72.2} & \textbf{71.1} & \textbf{79.8} & \textbf{74.8} & \textbf{81.8} & \textbf{76.2} \\
Full  & 69.5 & 62.7 & 78.3 & 70.5 & 79.7 & 73.8 \\
\bottomrule
\end{tabular}
}
\vspace{-5pt}
\end{table}

\subsection{Natural Sparsity and Layer-wise Replaceability}
\label{sec:exp_natural_sparse}

To examine how sparsity emerges in pretrained token mixing, we first construct dense replacement models by replacing all attention layers in DeiT baselines under the unified framework built in Sec.~\ref{sec:method_replacement}. 
These dense models provide a common basis for comparing token interaction patterns across the original attention backbone and different sequential substitutes.
In this setting, full replacement is already effective at small scale, while a clearer performance gap appears as model size increases; detailed results across scales and transfer benchmarks are provided in Appendix~\ref{sec:app_full_replace} (see supplementary materials).

We then use these dense models to characterize \emph{natural sparsity} in token interactions.
Figure~\ref{fig:natural_sparse} visualizes representative layers of DeiT and its fully replaced Mamba- and LSTM-based counterparts.
For DeiT, token interactions are shown by attention-score maps.
For replacements, which do not expose attention matrices, we use gradient-based interaction maps to quantify how strongly each output token depends on each input token.
Despite the different mechanisms, the maps reveal a coherent depth-dependent evolution in token mixing.

In shallow layers, the interaction patterns are relatively dense and spread over broad regions.
They also vary across architectures, suggesting that early token mixing is both global and model-dependent, and that different substitutes may realize this global mixing in different ways.
In deeper layers, the interaction patterns become markedly sparser and more concentrated on a small subset of tokens.
At the same time, the overall structures across DeiT, Mamba, and LSTM become more similar, indicating that the token-mixing behavior in later layers is less about globally redistributing information and more about selectively aggregating a few salient signals.
Taken together, these observations suggest that token mixing becomes effectively simpler and more selective with depth, and that the deepest layers are natural candidates for successful replacement.

To test whether this natural sparsity is predictive of replacement difficulty, we perform controlled layer-group replacements.
We replace one contiguous layer group at a time while keeping the remaining layers unchanged, and, for comparability, we update only the parameters of the substituted layers during training.
All rows in Table~\ref{tab:layer_group_replace} use the same training budget (50 epochs) and the same schedule, so performance differences reflect replacement difficulty rather than optimization budget; complete settings are provided in Appendix~\ref{sec:app_setup}.
A consistent ordering emerges across scales and substitute architectures: replacing later layers yields better performance than replacing earlier layers, and the deepest group is the easiest to replace.
For reference, we additionally report a fully replaced model trained to convergence with global finetuning in Appendix~\ref{sec:app_full_replace}.
Combined with Fig.~\ref{fig:natural_sparse}, this indicates that naturally sparser layers are also the ones that are more replaceable by sequential token mixers.

These results provide the first part of our sparsity-to-replaceability argument.
Natural layer-wise sparsity is not merely an incidental artifact, but already signals where attention replacement is most effective.
In the next subsection, we make sparsity explicit and controllable, and examine whether increasing sparsity further improves replaceability.

\begin{figure}[t]
    \centering
    \begin{subfigure}[t]{0.48\linewidth}
        \centering
        \includegraphics[width=\linewidth]{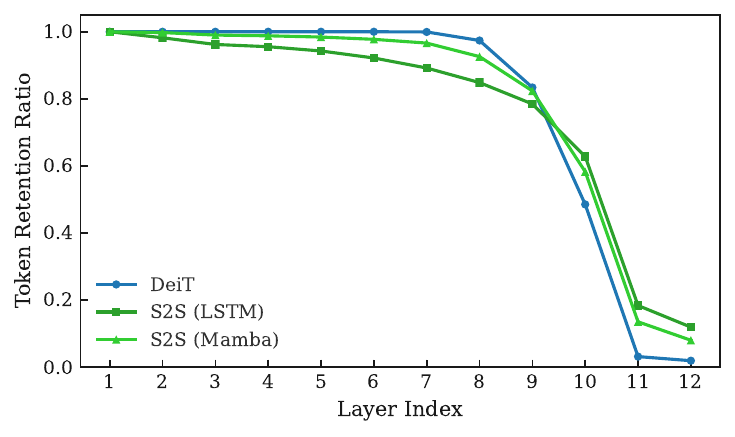}
        \caption{\textbf{Layer-wise token retention under A-ViT.}
        Average token retention ratio across layers for Attention-, Mamba- and LSTM-based replacement models after introducing A-ViT.}
        \label{fig:avit_retention}
    \end{subfigure}
    \hfill
    \begin{subfigure}[t]{0.50\linewidth}
        \centering
        \includegraphics[width=\linewidth]{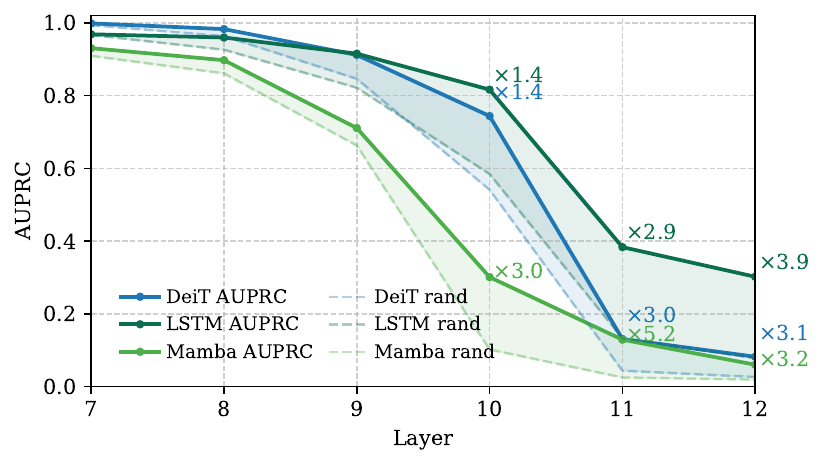}
        \caption{\textbf{Correlation between A-ViT-kept tokens and token importance.}
        AUPRC between the tokens retained by A-ViT and token importance, compared with random selection.}
        \label{fig:avit_importance}
    \end{subfigure}
    \caption{\textbf{Explicit sparsity induced by A-ViT.}
    Layer-wise token retention and its relation to token importance.}
    \label{fig:avit_sparse}
\end{figure}

\begin{figure}[!t]
    \centering
    \includegraphics[width=0.9\linewidth]{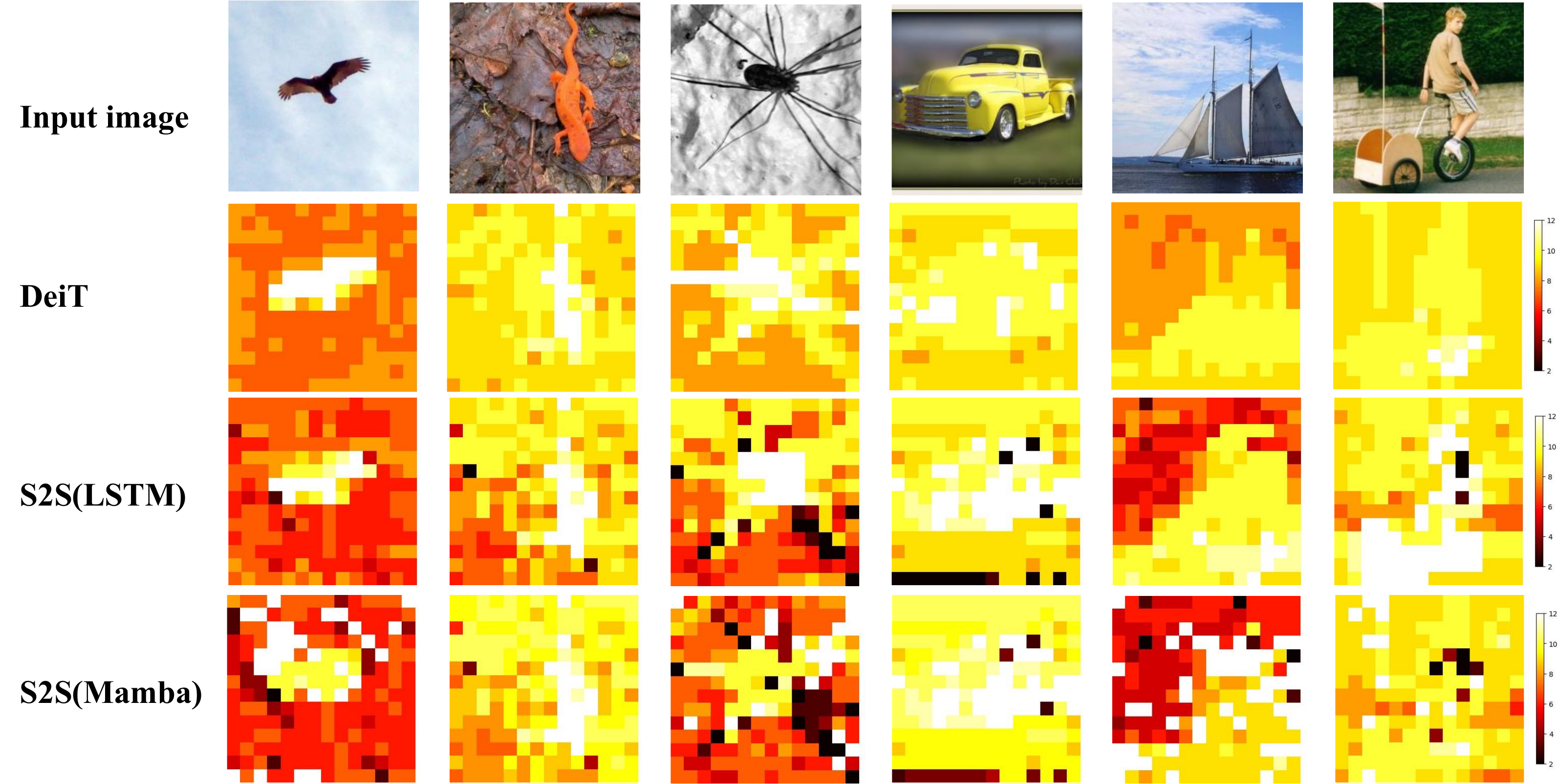}
    \caption{
        \textbf{Sample-level token retention visualization under A-ViT.}
        Token retention depth for representative ImageNet samples, shown for the A-ViT teacher and the corresponding LSTM- and Mamba-based replacement models.
        Brighter regions indicate tokens retained through more layers.}
    \label{fig:token_retention_vis}
\end{figure}

\subsection{Explicit Sparsity and Replacement Difficulty}
\label{sec:exp_explicit_sparse}

The results in Sec.~\ref{sec:exp_natural_sparse} show that naturally sparse layers are also the ones that are easier to replace. 
However, natural sparsity is an observed property of pretrained models and cannot be directly controlled.
To make this layer-wise trend explicit and adjustable, we introduce A-ViT as a token-retention mechanism and study sparsity under the sparsity-guided distillation framework of Sec.~\ref{sec:method_sparsity}.

As in the dense setting, we begin from fully replaced models so that the teacher and different sequential substitutes can be compared under the same replacement scope.
Specifically, we construct sparse replacement models by distilling from a pretrained A-ViT teacher into fully replaced Mamba- and LSTM-based students following Sec.~\ref{sec:method_sparsity}.
This gives a controlled setting in which sparsity is imposed through the same layer-wise masking mechanism, while the token-mixing operator is varied through the choice of substitute architecture.

We first examine how this explicit sparsity is distributed across layers.
Figure~\ref{fig:avit_sparse} statistically characterizes the masks induced by A-ViT.
Figure~\ref{fig:avit_retention} shows that the retention ratio remains close to one in shallow layers, but drops sharply in deeper layers, showing that explicit sparsity is highly uneven across depth rather than uniformly applied.
This profile closely matches the natural trend observed in Sec.~\ref{sec:exp_natural_sparse}: early layers still preserve most tokens, whereas later layers can sustain much more aggressive pruning.
At the same time, Figure~\ref{fig:avit_importance} compares the tokens selected by A-ViT with token importance estimates and reports their correlation against random selection.
Across the deeper layers, the A-ViT-kept tokens show consistently stronger alignment with important tokens than random baselines, indicating that the learned masks preserve informative tokens instead of imposing arbitrary sparsity.

Figure~\ref{fig:token_retention_vis} provides a sample-level view of the same phenomenon.
Across representative ImageNet samples, the A-ViT teacher and the corresponding sequential students exhibit similar hierarchical token-retention patterns: broad retention in earlier processing stages and progressively more selective retention in later ones.
This qualitative visualization complements the statistics in Fig.~\ref{fig:avit_sparse} and shows that the induced sparsity is not only measurable on average, but also visually consistent across models on individual examples.

\begin{figure}[t]
\centering
\includegraphics[width=0.6\linewidth]{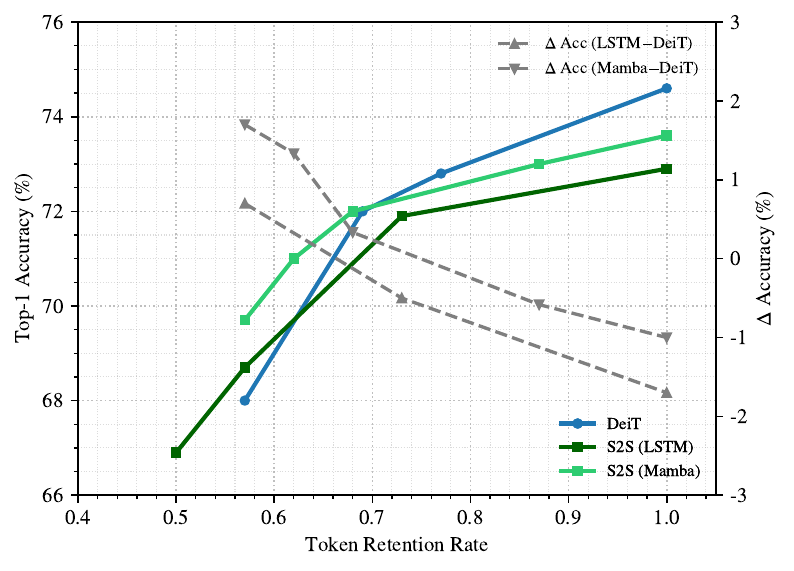}
\caption{\textbf{Accuracy under controlled token retention.}
ImageNet top-1 accuracy of the A-ViT teacher and the fully replaced sequential students at different token retention ratios.
Dashed lines show the accuracy gap relative to the teacher. 
As retention decreases, the gap between teacher and students consistently becomes smaller.}
\label{fig:retention_acc}
\end{figure}

We next ask whether stronger explicit sparsity also simplifies replacement.
To isolate this, we vary the token retention ratio and evaluate the resulting teacher and student accuracy under full replacement.
Figure~\ref{fig:retention_acc} shows the corresponding curves for the Tiny-scale setting.
As the retention ratio decreases, the student--teacher gap consistently shrinks for both replacements.
This indicates that explicit sparsification reproduces the layer-wise sparsity profile of the teacher, as well as reduces the difficulty of approximating its token-mixing behavior.

These results provide the second half of our sparsity-to-replaceability argument.
While Sec.~\ref{sec:exp_natural_sparse} shows that natural sparsity predicts where replacement is easier, the present section shows that explicit sparsity can further improve replaceability in a controlled manner.
In the next subsection, we combine these two observations and show that applying sparsity-guided replacement to naturally sparse layers yields a favorable accuracy--efficiency tradeoff.

\begin{table}[t]
\centering
\caption{\textbf{Accuracy--efficiency tradeoff on DeiT-Small under matched sparsity budgets.}
ImageNet top-1 accuracy (\%) and token-mixing throughput (tokens/ms) at $224^2$ and $384^2$ resolution.
Throughput is measured on the token-mixing component, and model speedup is estimated from the measured token-mixer runtime while keeping the remaining backbone unchanged.}
\label{tab:tradeoff}
{
\setlength{\tabcolsep}{5pt}
\renewcommand{\arraystretch}{1.08}
\begin{tabular}{lccccccc}
\toprule
\textbf{Model} & \textbf{Layers} & \textbf{Ret.} & \textbf{Top-1} 
& \multicolumn{2}{c}{\makecell{\textbf{Measured token}\\ \textbf{throughput} $\uparrow$}} 
& \multicolumn{2}{c}{\makecell{\textbf{Estimated model}\\ \textbf{speedup} $\uparrow$}} \\
\cmidrule(lr){5-6} \cmidrule(lr){7-8}
& & & 
& \textbf{Cls $224^2$} & \textbf{Cls $384^2$} 
& \textbf{Cls $224^2$} & \textbf{Cls $384^2$} \\
\midrule
DeiT & None & 1.00 & 79.8 & 12.36 & 3.12 & 1.00$\times$ & 1.00$\times$ \\
\midrule
\multirow{2}{*}{A-ViT}
& \multirow{2}{*}{None} & 0.78 & 78.6 & 13.51 & 4.08 & 1.09$\times$ & 1.31$\times$  \\
&  & 0.70 & 78.2 & 15.03 & 4.71 & 1.22$\times$ & 1.51$\times$  \\
\midrule
\multirow{3}{*}{S2S-Mamba}
& \multirow{3}{*}{9--11} & 1.00 & 79.3 & 10.46 & 3.81 & 0.85$\times$ & 1.22$\times$ \\
&  & 0.78 & \textbf{78.6} & 13.49 & \textbf{5.11} & 1.09$\times$ & \textbf{1.64$\times$} \\
&  & 0.70 & \textbf{78.4} & 14.45 & \textbf{5.32} & 1.17$\times$ & \textbf{1.71$\times$} \\
\bottomrule
\end{tabular}
}
\end{table}
\subsection{Layer-aware Sparse Replacement for Better Tradeoff}
\label{sec:tradeoff}

The results in Secs.~\ref{sec:exp_natural_sparse} and \ref{sec:exp_explicit_sparse} together suggest that sparsity should not be applied uniformly across the network.
Natural sparsity reveals that the deepest layers are already the most selective in token mixing and therefore the most replaceable, while explicit sparsity further reduces the difficulty of learning the corresponding target mappings.
Taken together, these observations motivate a layer-aware sparse replacement strategy that concentrates replacement on the naturally sparse layers and trains the replacement under controlled sparsity.

Table~\ref{tab:tradeoff} evaluates this strategy on DeiT-Small under matched token-retention ratios.
Dense DeiT serves as the dense reference, and A-ViT serves as the sparse attention baseline that preserves attention as the token-mixing operator.
The proposed model instead replaces only the deepest layer group identified in Sec.~\ref{sec:exp_natural_sparse} and trains this replacement with the sparsity-guided distillation framework of Sec.~\ref{sec:method_sparsity}.
Under this design, the sparsity budget remains comparable across models, while the key difference lies in whether sparsity is used to retain sparse attention throughout the network or to guide replacement on the layers that are already most amenable to substitution.

The results show that the proposed strategy uses the sparsity budget more effectively.
At comparable retention ratios, the layer-aware sparse replacement reaches, and in some settings slightly exceeds, the accuracy of the corresponding sparse attention baseline.
This improvement is not obtained by simply pruning more aggressively, nor by replacing more layers.
Rather, it arises from aligning replacement with the layer-wise sparsity structure of the pretrained model: the deepest layers are already the ones whose token mixing is the most selective, so replacing them incurs less functional mismatch, while explicit sparsity further simplifies the target mapping that the sequential substitute must learn.

The same pattern appears on the efficiency side.
Table~\ref{tab:tradeoff} reports both the measured token-mixing throughput (tokens/ms) and the corresponding estimated model-level speedup.
At the same retention ratio, sparse attention can be competitive at smaller token counts, whereas the advantage of layer-aware sparse replacement becomes clear at larger token counts.
In particular, the proposed replacement achieves higher token-mixing throughput and larger estimated speedup at $384^2$ resolution, while maintaining comparable accuracy under the same sparsity budget.
This indicates that the proposed strategy improves not only the approximation side of the problem, but also the computational side: once sparsity is concentrated on the replaceable layers, the sequential substitute can turn the same token budget into more favorable runtime behavior.

Taken together, these results close the loop of our sparsity-to-simplicity argument.
Natural layer-wise sparsity predicts where attention is easier to replace, and explicit sparsification further reduces the student--teacher gap under controlled retention.
Combining the two yields a replacement strategy that uses a comparable sparsity budget more effectively, achieving a more favorable accuracy--efficiency operating point than sparse attention baselines.
\section{Conclusion}
\label{sec:conclusion}

We revisit attention replacement from the perspective of sparsity. 
Our study shows that sparsity is strongly layer-dependent in pretrained visual token mixing, and that this variation is closely related to replaceability: naturally sparse layers are consistently easier to replace with sequential token mixers than dense ones. 
We further show that explicit sparsification via A-ViT strengthens this effect, as increasing sparsity consistently reduces the student--teacher gap for both BiLSTM and BiMamba replacements. 

Based on these observations, we combine natural and explicit sparsity in a layer-aware replacement strategy, where sequential replacement is focused on the most replaceable layers and trained under controlled sparsity. 
This yields a better accuracy--efficiency tradeoff than sparse attention baselines under comparable sparsity, showing that sparsity is not only an efficiency property of pretrained attention, but also a useful signal for where replacement should occur and how it should be learned.

Overall, our results suggest that attention replacement should be treated as a structured, layer-dependent problem rather than a uniform architecture swap.
We hope this perspective will encourage further study of replacement strategies that combine pretrained transformer knowledge with simpler and more efficient sequence modules.


%
%
\bibliographystyle{splncs04}
\bibliography{main}

@String(CVPR  = {IEEE Conf. Comput. Vis. Pattern Recog.})

@String(ICML  = {Int. Conf. Mach. Learn.})

@String(ICLR  = {Int. Conf. Learn. Represent.})

@String(CVPR  = {CVPR})

@String(ICML  = {ICML})

@String(ICLR  = {ICLR})

@inproceedings{vaswani2023attention,
 author = {Vaswani, Ashish and Shazeer, Noam and Parmar, Niki and Uszkoreit, Jakob and Jones, Llion and Gomez, Aidan N and Kaiser, \L ukasz and Polosukhin, Illia},
 booktitle = {Advances in Neural Information Processing Systems},
 title = {Attention is All you Need},
 year = {2017}
}

@inproceedings{devlin2019bert,
    title = "{BERT}: Pre-training of Deep Bidirectional Transformers for Language Understanding",
    author = "Devlin, Jacob  and
      Chang, Ming-Wei  and
      Lee, Kenton  and
      Toutanova, Kristina",
    booktitle = "Proceedings of the 2019 Conference of the North {A}merican Chapter of the Association for Computational Linguistics: Human Language Technologies, Volume 1 (Long and Short Papers)",
    year = "2019",
    publisher = "Association for Computational Linguistics",
    url = "https://aclanthology.org/N19-1423/",
    doi = "10.18653/v1/N19-1423",
    pages = "4171--4186",
}

@inproceedings{
xiao2024streamingllm,
title={Efficient Streaming Language Models with Attention Sinks},
author={Guangxuan Xiao and Yuandong Tian and Beidi Chen and Song Han and Mike Lewis},
booktitle={The Twelfth International Conference on Learning Representations},
year={2024},
url={https://openreview.net/forum?id=NG7sS51zVF}
}

@inproceedings{
zhang2023ho,
title={H2O: Heavy-Hitter Oracle for Efficient Generative Inference of Large Language Models},
author={Zhenyu Zhang and Ying Sheng and Tianyi Zhou and Tianlong Chen and Lianmin Zheng and Ruisi Cai and Zhao Song and Yuandong Tian and Christopher Re and Clark Barrett and Zhangyang Wang and Beidi Chen},
booktitle={Thirty-seventh Conference on Neural Information Processing Systems},
year={2023},
url={https://openreview.net/forum?id=RkRrPp7GKO}
}

@misc{
cai2025pyramidkv,
title={Pyramid{KV}: Dynamic {KV} Cache Compression based on Pyramidal Information Funneling},
author={Zefan Cai and Yichi Zhang and Bofei Gao and Yuliang Liu and Tianyu Liu and Keming Lu and Wayne Xiong and Yue Dong and Junjie Hu and Wen Xiao},
year={2025},
url={https://openreview.net/forum?id=jZVNmDiU86}
}

@inproceedings{
li2024snapkv,
title={Snap{KV}: {LLM} Knows What You are Looking for Before Generation},
author={Yuhong Li and Yingbing Huang and Bowen Yang and Bharat Venkitesh and Acyr Locatelli and Hanchen Ye and Tianle Cai and Patrick Lewis and Deming Chen},
booktitle={The Thirty-eighth Annual Conference on Neural Information Processing Systems},
year={2024},
url={https://openreview.net/forum?id=poE54GOq2l}
}

@inproceedings{dosovitskiy2021vit,
  author       = "Alexey Dosovitskiy and
                  Lucas Beyer and
                  Alexander Kolesnikov and
                  Dirk Weissenborn and
                  Xiaohua Zhai and
                  Thomas Unterthiner and
                  Mostafa Dehghani and
                  Matthias Minderer and
                  Georg Heigold and
                  Sylvain Gelly and
                  Jakob Uszkoreit and
                  Neil Houlsby",
  title        = "An Image is Worth 16x16 Words: Transformers for Image Recognition
                  at Scale",
  booktitle    = "9th International Conference on Learning Representations, {ICLR} 2021,
                  Virtual Event, Austria, May 3-7, 2021",
  year         = "2021",
}

@inproceedings{radford2021clip,
  author       = {Alec Radford and
                  Jong Wook Kim and
                  Chris Hallacy and
                  Aditya Ramesh and
                  Gabriel Goh and
                  Sandhini Agarwal and
                  Girish Sastry and
                  Amanda Askell and
                  Pamela Mishkin and
                  Jack Clark and
                  Gretchen Krueger and
                  Ilya Sutskever},
  editor       = {Marina Meila and
                  Tong Zhang},
  title        = {Learning Transferable Visual Models From Natural Language Supervision},
  booktitle    = {Proceedings of the 38th International Conference on Machine Learning,
                  {ICML} 2021, 18-24 July 2021, Virtual Event},
  series       = {Proceedings of Machine Learning Research},
  volume       = {139},
  pages        = {8748--8763},
  year         = {2021},
  url          = {http://proceedings.mlr.press/v139/radford21a.html},
}

@article{tay2022efficient,
author = {Tay, Yi and Dehghani, Mostafa and Bahri, Dara and Metzler, Donald},
title = {Efficient Transformers: A Survey},
year = {2022},
issue_date = {June 2023},
publisher = {Association for Computing Machinery},
volume = {55},
number = {6},
issn = {0360-0300},
url = {https://doi.org/10.1145/3530811},
doi = {10.1145/3530811},
journal = {ACM Comput. Surv.},
month = dec,
articleno = {109},
numpages = {28}
}

@inproceedings{dao2022flash,
author = {Dao, Tri and Fu, Daniel Y. and Ermon, Stefano and Rudra, Atri and R\'{e}, Christopher},
title = {FLASHATTENTION: fast and memory-efficient exact attention with IO-awareness},
year = {2022},
isbn = {9781713871088},
publisher = {Curran Associates Inc.},
booktitle = {Proceedings of the 36th International Conference on Neural Information Processing Systems},
articleno = {1189},
numpages = {16},
series = {NIPS '22}
}

@misc{cordonnier2020relationship,
      title={On the Relationship between Self-Attention and Convolutional Layers}, 
      author={Jean-Baptiste Cordonnier and Andreas Loukas and Martin Jaggi},
      year={2020},
      eprint={1911.03584},
      archivePrefix={arXiv},
      primaryClass={cs.LG},
      url={https://arxiv.org/abs/1911.03584}, 
}

@inproceedings{tolstikhin2021mlpmixer,
author = {Tolstikhin, Ilya and Houlsby, Neil and Kolesnikov, Alexander and Beyer, Lucas and Zhai, Xiaohua and Unterthiner, Thomas and Yung, Jessica and Steiner, Andreas and Keysers, Daniel and Uszkoreit, Jakob and Lucic, Mario and Dosovitskiy, Alexey},
title = {MLP-mixer: an all-MLP architecture for vision},
year = {2021},
isbn = {9781713845393},
publisher = {Curran Associates Inc.},
booktitle = {Proceedings of the 35th International Conference on Neural Information Processing Systems},
articleno = {1857},
numpages = {12},
series = {NIPS '21}
}

@misc{
sun2023retnet,
title={Retentive Network: A Successor to Transformer for Large Language Models},
author={Yutao Sun and Li Dong and Shaohan Huang and Shuming Ma and Yuqing Xia and Jilong Xue and Jianyong Wang and Furu Wei},
year={2024},
url={https://openreview.net/forum?id=UU9Icwbhin}
}

@inproceedings{
gu2024mamba,
title={Mamba: Linear-Time Sequence Modeling with Selective State Spaces},
author={Albert Gu and Tri Dao},
booktitle={First Conference on Language Modeling},
year={2024},
url={https://openreview.net/forum?id=tEYskw1VY2}
}

@misc{
he2024matterstransformers,
title={What Matters in Transformers? Not All Attention is Needed},
author={Shwai He and Guoheng Sun and Zheyu Shen and Ang Li},
year={2025},
url={https://openreview.net/forum?id=YLTWwEjkdx}
}

@misc{bhojanapalli2021leveragingredundancy,
      title={Leveraging redundancy in attention with Reuse Transformers}, 
      author={Srinadh Bhojanapalli and Ayan Chakrabarti and Andreas Veit and Michal Lukasik and Himanshu Jain and Frederick Liu and Yin-Wen Chang and Sanjiv Kumar},
      year={2021},
      eprint={2110.06821},
      archivePrefix={arXiv},
      primaryClass={cs.LG},
      url={https://arxiv.org/abs/2110.06821}, 
}

@InProceedings{touvron2021deit,
  title = 	 {Training data-efficient image transformers \& distillation through attention},
  author =       {Touvron, Hugo and Cord, Matthieu and Douze, Matthijs and Massa, Francisco and Sablayrolles, Alexandre and Jegou, Herve},
  booktitle = 	 {Proceedings of the 38th International Conference on Machine Learning},
  pages = 	 {10347--10357},
  year = 	 {2021},
  editor = 	 {Meila, Marina and Zhang, Tong},
  volume = 	 {139},
  series = 	 {Proceedings of Machine Learning Research},
  month = 	 {18--24 Jul},
  publisher =    {PMLR},
  url = 	 {https://proceedings.mlr.press/v139/touvron21a.html},
}

@ARTICLE{hochreiter1997long,
  author={Hochreiter, Sepp and Schmidhuber, Jürgen},
  journal={Neural Computation}, 
  title={Long Short-Term Memory}, 
  year={1997},
  volume={9},
  number={8},
  pages={1735-1780},
  doi={10.1162/neco.1997.9.8.1735}}

@article{hinton2015distilling,
  title={Distilling the Knowledge in a Neural Network},
  author={Geoffrey E. Hinton and Oriol Vinyals and Jeffrey Dean},
  journal={ArXiv},
  year={2015},
  volume={abs/1503.02531},
}

@inproceedings{jiao2020tinybert,
    title = "{T}iny{BERT}: Distilling {BERT} for Natural Language Understanding",
    author = "Jiao, Xiaoqi  and
      Yin, Yichun  and
      Shang, Lifeng  and
      Jiang, Xin  and
      Chen, Xiao  and
      Li, Linlin  and
      Wang, Fang  and
      Liu, Qun",
    editor = "Cohn, Trevor  and
      He, Yulan  and
      Liu, Yang",
    booktitle = "Findings of the Association for Computational Linguistics: EMNLP 2020",
    month = nov,
    year = "2020",
    publisher = "Association for Computational Linguistics",
    url = "https://aclanthology.org/2020.findings-emnlp.372/",
    doi = "10.18653/v1/2020.findings-emnlp.372",
    pages = "4163--4174",
}

@inproceedings{sun2020mobilebert,
    title = "{M}obile{BERT}: a Compact Task-Agnostic {BERT} for Resource-Limited Devices",
    author = "Sun, Zhiqing  and
      Yu, Hongkun  and
      Song, Xiaodan  and
      Liu, Renjie  and
      Yang, Yiming  and
      Zhou, Denny",
    editor = "Jurafsky, Dan  and
      Chai, Joyce  and
      Schluter, Natalie  and
      Tetreault, Joel",
    booktitle = "Proceedings of the 58th Annual Meeting of the Association for Computational Linguistics",
    month = jul,
    year = "2020",
    publisher = "Association for Computational Linguistics",
    url = "https://aclanthology.org/2020.acl-main.195/",
    doi = "10.18653/v1/2020.acl-main.195",
    pages = "2158--2170",
}

@inproceedings{
fan2019reducing,
title={Reducing Transformer Depth on Demand with Structured Dropout},
author={Angela Fan and Edouard Grave and Armand Joulin},
booktitle={International Conference on Learning Representations},
year={2020},
url={https://openreview.net/forum?id=SylO2yStDr}
}

@inproceedings{wang2020minilm,
author = {Wang, Wenhui and Wei, Furu and Dong, Li and Bao, Hangbo and Yang, Nan and Zhou, Ming},
title = {MINILM: deep self-attention distillation for task-agnostic compression of pre-trained transformers},
year = {2020},
isbn = {9781713829546},
publisher = {Curran Associates Inc.},
booktitle = {Proceedings of the 34th International Conference on Neural Information Processing Systems},
articleno = {485},
numpages = {13},
series = {NIPS '20}
}

@techreport{krizhevsky2009learning,
  title={Learning Multiple Layers of Features from Tiny Images},
  author={Krizhevsky, Alex},
  year={2009},
  institution={University of Toronto},
  url={https://www.cs.toronto.edu/~kriz/learning-features-2009-TR.pdf}
}

@inproceedings{krause20133d,
  title={3D Object Representations for Fine-Grained Categorization},
  author={Krause, Jonathan and Stark, Michael and Deng, Jia and Fei-Fei, Li},
  booktitle={Proceedings of the IEEE Conference on Computer Vision and Pattern Recognition (CVPR)},
  year={2013},
  pages={554--561}
}

@inproceedings{nilsback2008automated,
  title={Automated flower classification over a large number of classes},
  author={Nilsback, M-E and Zisserman, A},
  booktitle={Proceedings of the Indian Conference on Computer Vision, Graphics and Image Processing},
  year={2008},
  pages={722--729}
}

@inproceedings{vanhorn2018inaturalist,
  title={The iNaturalist species classification and detection dataset},
  author={Van Horn, Grant and Mac Aodha, Oisin and Song, Yang and Cui, Chenyi and Sun, Yin and Shepard, Andrew and Adam, Hartwig and Perona, Pietro and Belongie, Serge},
  booktitle={Proceedings of the IEEE Conference on Computer Vision and Pattern Recognition (CVPR)},
  year={2018},
  pages={8769--8778}
}

@inproceedings{katharopoulos2020transformers,
  title={Transformers are rnns: Fast autoregressive transformers with linear attention},
  author={Katharopoulos, Angelos and Vyas, Apoorv and Pappas, Nikolaos and Fleuret, Fran{\c{c}}ois},
  booktitle={International conference on machine learning},
  pages={5156--5165},
  year={2020},
  organization={PMLR}
}

@inproceedings{NEURIPS2024_mambainllama,
  title = {The Mamba in the Llama: Distilling and Accelerating Hybrid Models},
  author = {Wang, Junxiong and Paliotta, Daniele and May, Avner and Rush, Alexander M. and Dao, Tri},
  booktitle = {Advances in Neural Information Processing Systems},
  editor = {A. Globerson and L. Mackey and D. Belgrave and A. Fan and U. Paquet and J. Tomczak and C. Zhang},
  pages = {62432--62457},
  publisher = {Curran Associates, Inc.},
  volume = {37},
  year = {2024}
}

@inproceedings{
wen2025rnns,
title={{RNN}s are not Transformers (Yet):  The Key Bottleneck on In-Context Retrieval},
author={Kaiyue Wen and Xingyu Dang and Kaifeng Lyu},
booktitle={The Thirteenth International Conference on Learning Representations},
year={2025},
url={https://openreview.net/forum?id=h3wbI8Uk1Z}
}

@inproceedings{DynamicViT2021,
author = {Rao, Yongming and Zhao, Wenliang and Liu, Benlin and Lu, Jiwen and Zhou, Jie and Hsieh, Cho-Jui},
title = {DynamicViT: efficient vision transformers with dynamic token sparsification},
year = {2021},
isbn = {9781713845393},
address = {Red Hook, NY, USA},
booktitle = {Proceedings of the 35th International Conference on Neural Information Processing Systems},
articleno = {1068},
numpages = {13},
series = {NIPS '21}
}

@inproceedings{TokenLearner2021,
 author = {Ryoo, Michael and Piergiovanni, AJ and Arnab, Anurag and Dehghani, Mostafa and Angelova, Anelia},
 booktitle = {Advances in Neural Information Processing Systems},
 pages = {12786--12797},
 title = {TokenLearner: Adaptive Space-Time Tokenization for Videos},
 url = {https://proceedings.neurips.cc/paper_files/paper/2021/file/6a30e32e56fce5cf381895dfe6ca7b6f-Paper.pdf},
 volume = {34},
 year = {2021}
}

@InProceedings{Yin_2022_CVPR,
    author    = {Yin, Hongxu and Vahdat, Arash and Alvarez, Jose M. and Mallya, Arun and Kautz, Jan and Molchanov, Pavlo},
    title     = {A-ViT: Adaptive Tokens for Efficient Vision Transformer},
    booktitle = {Proceedings of the IEEE/CVF Conference on Computer Vision and Pattern Recognition (CVPR)},
    month     = {June},
    year      = {2022},
    pages     = {10809-10818}
}
\clearpage
\appendix\section{Full Replacement Baseline}
\label{sec:app_full_replace}

For completeness, we report the dense full-replacement results that provide the baseline context for the sparsity-driven analysis in the main paper.
In this setting, all attention layers are replaced by sequential token mixers under the interface-aligned design in Sec.~\ref{sec:method_replacement}.
These experiments verify that full attention replacement is feasible before introducing sparsity, and also show how replacement difficulty changes with model scale.
Unless otherwise specified, the fully replaced models are trained for 200 epochs with only the replacement layers updated, followed by 100 epochs of global finetuning.

\begin{table}[h]
    \centering
    \caption{
        \textbf{Full replacement results across model scales and classification benchmarks.}
        Top-1 accuracy (\%) of DeiT and its fully replaced sequential variants on ImageNet and transfer classification datasets, together with model size (M).
        }
    \label{tab:full_replace_results}
\resizebox{\textwidth}{!}{
\begin{tabular}{llcccccccc}
\toprule
\textbf{Scale} & \textbf{Model} & \textbf{Params} & \textbf{ImageNet} & \textbf{CIFAR-10} & \textbf{CIFAR-100} & \textbf{Cars} & \textbf{Flowers} & \textbf{iNat-18} & \textbf{iNat-19} \\
\midrule
\multirow{3}{*}{Tiny} & Attn & 5.7 & 72.2 & 97.9 & 85.7 & 90.5 & 97.4 & 62.4 & 72.1\\
& \cellcolor{orange!20}Mamba & \cellcolor{orange!20}7.7 & \cellcolor{orange!20}73.4 & \cellcolor{orange!20}98.1 & \cellcolor{orange!20}85.7 & \cellcolor{orange!20}92.9 & \cellcolor{orange!20}98.1 & \cellcolor{orange!20}64.9 & \cellcolor{orange!20}72.2 \\
& \cellcolor{orange!20}LSTM & \cellcolor{orange!20}7.7 & \cellcolor{orange!20}73.3 & \cellcolor{orange!20}98.1 & \cellcolor{orange!20}85.4 & \cellcolor{orange!20}92.3 & \cellcolor{orange!20}97.9 & \cellcolor{orange!20}64.7 & \cellcolor{orange!20}71.3\\
\midrule
\multirow{3}{*}{Small} & Attn & 22.9 & 79.8 & 98.5 & 87.1 & 91.7 & 98.1  & 66.8 & 74.2 \\
& \cellcolor{orange!20}Mamba & \cellcolor{orange!20}26.9 & \cellcolor{orange!20}79.3 & \cellcolor{orange!20}98.5 & \cellcolor{orange!20}87.1 & \cellcolor{orange!20}92.1 & \cellcolor{orange!20}98.1 & \cellcolor{orange!20}67.1 & \cellcolor{orange!20}74.3 \\
& \cellcolor{orange!20}LSTM & \cellcolor{orange!20}23.9 & \cellcolor{orange!20}77.8 & \cellcolor{orange!20}98.3 & \cellcolor{orange!20}86.8 & \cellcolor{orange!20}92.1 & \cellcolor{orange!20}97.9 & \cellcolor{orange!20}67.0 & \cellcolor{orange!20}73.9 \\
\midrule
\multirow{3}{*}{Base} & Attn & 86.5 & 81.8 & 99.1 & 90.8 & 92.1 & 98.4 & 73.2 & 77.7 \\
& \cellcolor{orange!20}Mamba & \cellcolor{orange!20}89.9 & \cellcolor{orange!20}80.5 & \cellcolor{orange!20}98.7 & \cellcolor{orange!20}89.4 & \cellcolor{orange!20}92.5 & \cellcolor{orange!20}98.1 & \cellcolor{orange!20}71.2 & \cellcolor{orange!20}76.9 \\
& \cellcolor{orange!20}LSTM & \cellcolor{orange!20}83.2 & \cellcolor{orange!20}79.7 & \cellcolor{orange!20}98.2 & \cellcolor{orange!20}87.7 & \cellcolor{orange!20}92.3 & \cellcolor{orange!20}97.9 & \cellcolor{orange!20}67.0 & \cellcolor{orange!20}75.1 \\
\bottomrule
\end{tabular}
}
\end{table}

Table~\ref{tab:full_replace_results} summarizes the results on ImageNet and the transfer classification benchmarks used in DeiT.
For DeiT-Tiny, full replacement is already highly effective: both BiMamba and BiLSTM match or slightly exceed the attention baseline on ImageNet, and remain competitive across the transfer datasets.
This indicates that, at small scale, the pretrained token-mixing behavior can be captured well by sequential substitutes under the proposed framework.

As the model scale increases, the replacement difficulty becomes more visible.
On DeiT-Small, BiMamba remains only slightly below the original attention model, while BiLSTM exhibits a clearer performance gap.
On DeiT-Base, the gap further enlarges, indicating that full replacement is no longer uniformly easy at larger scales.
These results provide the dense baseline underlying the sparsity-oriented study in the main paper: sequential replacement is feasible, but its difficulty depends strongly on scale and, as shown in the main text, on layer-wise sparsity.

\begin{figure}[t]
\centering
\includegraphics[width=0.8\linewidth]{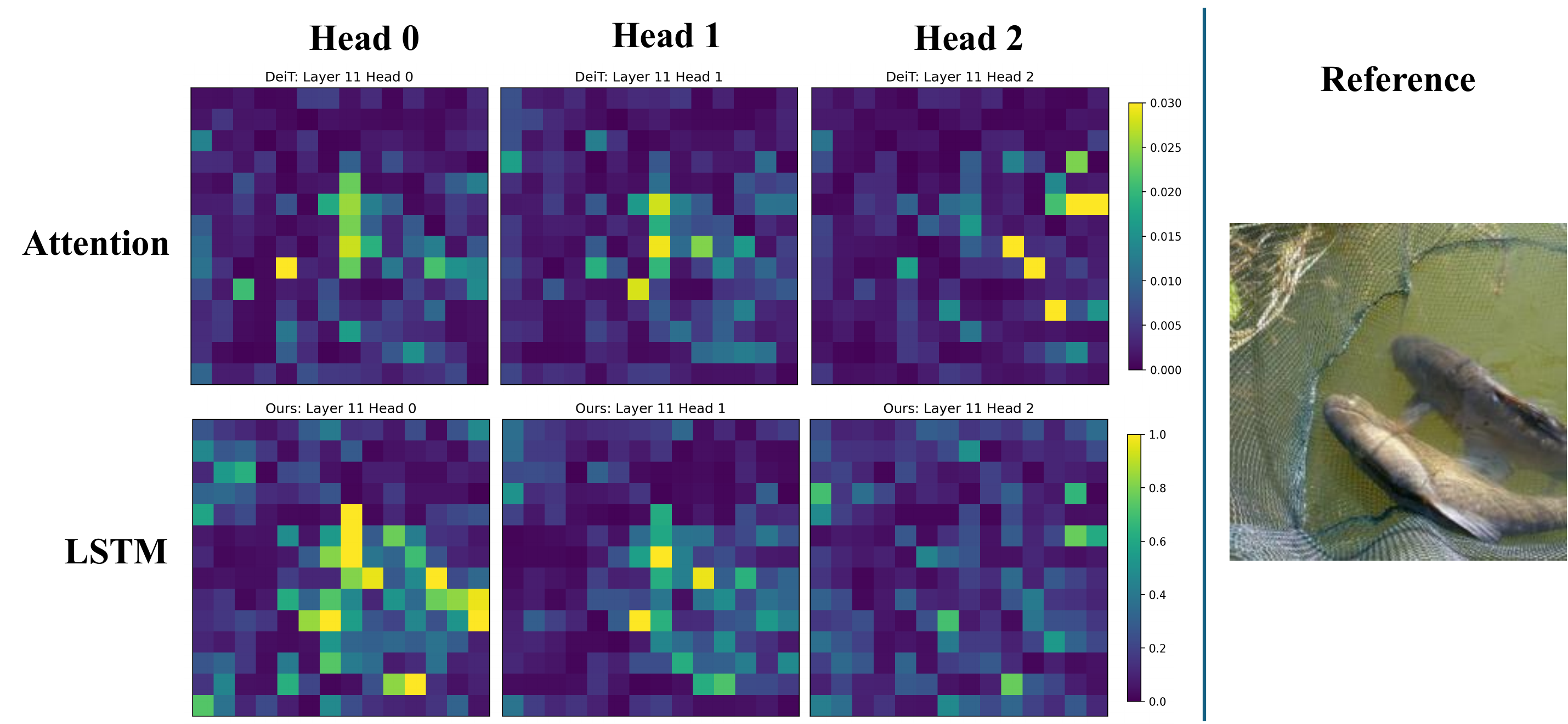}
\caption{\textbf{Head-wise token-interaction patterns.}
Representative head-wise interaction maps from attention- and LSTM-based models at the last layer.
Different heads show distinct interaction structures in both models.}
\label{fig:headwise_vis}
\end{figure}

To further test whether the same behavior transfers beyond ImageNet, we evaluate the ImageNet-trained checkpoints on the standard DeiT transfer benchmarks, including CIFAR-10/100~\cite{krizhevsky2009learning} for generic object classification, Stanford Cars~\cite{krause20133d} and Flowers-102~\cite{nilsback2008automated} for fine-grained recognition, and iNat-18/19~\cite{vanhorn2018inaturalist} for more challenging long-tailed natural category distributions.
Using the same benchmark suite as DeiT keeps the comparison directly aligned with prior work, while the diversity of these datasets provides a check of generalization across different classification regimes.
As shown in Table~\ref{tab:full_replace_results}, the overall trend is consistent with ImageNet: full replacement remains effective at smaller scales, while the gap grows as the backbone becomes larger and the tasks become more demanding.

To qualitatively examine the effect of the multi-head replacement design, Fig.~\ref{fig:headwise_vis} compares representative head-wise interaction patterns between the DeiT teacher and the corresponding dense replacement model.
Different heads capture distinct token interaction structures, and the sequential replacement preserves these differences at the head level.
This visualization provides qualitative support for retaining a multi-head interface in the sequential token mixer.

\section{Detailed Experimental Setups}
\label{sec:app_setup}

This section summarizes the experimental settings used in the main paper and the appendix.
Since different parts of the paper serve different purposes, we separate shared settings from the protocols specific to dense full replacement, sparsity-guided distillation, and tradeoff evaluation.

\subsection{Common Setup}
\label{sec:app_setup_common}

\paragraph{Datasets and backbones.}
Unless otherwise specified, all experiments are conducted on ImageNet with DeiT backbones~\cite{touvron2021deit}.
The dense full-replacement results in Appendix~\ref{sec:app_full_replace} additionally include the standard DeiT transfer classification benchmarks, namely CIFAR-10/100~\cite{krizhevsky2009learning}, Stanford Cars~\cite{krause20133d}, Flowers-102~\cite{nilsback2008automated}, and iNat-18/19~\cite{vanhorn2018inaturalist}.
Our main analysis focuses on DeiT-Tiny and DeiT-Small, while DeiT-Base is included when available for dense baselines and layer-group replacement.

\paragraph{Teacher and student models.}
The teacher is either a pretrained DeiT model or a pretrained A-ViT model, depending on the experiment.
The student follows the replacement framework in Sec.~\ref{sec:method_replacement}, where selected attention layers are replaced by sequential token mixers under the same layer interface.
Unless otherwise specified, the sequential substitutes are instantiated as BiLSTM or BiMamba under the shared multi-head bidirectional wrapper described in Sec.~\ref{sec:method_replacement}.

\paragraph{Optimization.}
We follow DeiT-style optimization with AdamW and cosine learning rate decay.
All experiments are implemented in PyTorch and run on NVIDIA A6000 GPUs.
Mixup and CutMix are disabled in the sparsity-guided distillation experiments, since token-level mixing augmentation interferes with halting supervision and layer-wise token masks.

\paragraph{A-ViT teachers.}
Since official A-ViT checkpoints are unavailable, we retrain A-ViT teachers from pretrained DeiT backbones.
The resulting teachers preserve competitive classification accuracy while producing stable and controllable layer-wise token-retention patterns, which are then used in Secs.~4.2 and~4.3.

\subsection{Setup for Appendix~\ref{sec:app_full_replace}: Dense Full Replacement}
\label{sec:app_setup_full}

For the dense full-replacement baselines reported in Appendix~\ref{sec:app_full_replace}, all attention layers are replaced under the framework of Sec.~\ref{sec:method_replacement}.
The student is first trained for 200 epochs with only the replacement layers updated, followed by 100 epochs of global finetuning.
This schedule is used only for the full-replacement baselines, whose goal is to report the converged performance of dense replacement across model scales and transfer benchmarks.

\subsection{Setup for Sec.~4.1: Natural Sparsity and Layer-wise Replaceability}
\label{sec:app_setup_41}

\paragraph{Dense replacement models for visualization.}
To visualize natural sparsity across architectures, we first construct dense replacement models by replacing all attention layers under the framework of Sec.~\ref{sec:method_replacement}.
These models are used only to analyze token-interaction patterns across the teacher and sequential substitutes.

\paragraph{Layer-group replacement ablations.}
For the controlled layer-group ablations in Sec.~4.1, one contiguous layer group is replaced at a time while all remaining layers are kept unchanged.
To ensure comparability across depths, all rows in Table~\ref{tab:layer_group_replace} use the same training budget and the same optimization schedule, and only the substituted layers are updated during training.
In our experiments, this ablation budget is fixed to 50 epochs.
This design isolates the replaceability of the selected layers from changes elsewhere in the backbone.

\subsection{Setup for Sec.~4.2: Explicit Sparsity and Replacement Difficulty}
\label{sec:app_setup_42}

For Sec.~4.2, we construct sparse replacement models by distilling from pretrained A-ViT teachers into fully replaced BiLSTM- and BiMamba-based students following Sec.~\ref{sec:method_sparsity}.
All results in Sec.~4.2 use the two-stage sparsity-guided distillation procedure.
Stage~1 is trained for 200 epochs with learning rate $5\times 10^{-4}$ under teacher masks, and Stage~2 is trained for 100 epochs with learning rate $5\times 10^{-5}$ using the student’s own halting modules.
Unless otherwise specified, the loss coefficients are fixed as
$\lambda_{\mathrm{cls}}=1.0$,
$\lambda_{\mathrm{sim}}=0.75$,
and
$\lambda_{\mathrm{avit}}=\lambda_{\mathrm{mask}}=0.1$.

\subsection{Setup for Sec.~4.3: Layer-aware Sparse Replacement}
\label{sec:app_setup_43}

Based on the observations in Sec.~4.1, Sec.~4.3 focuses on the deepest layer group, which is both the most naturally sparse and the most replaceable.
In our DeiT-Small experiments, this corresponds to replacing layers 9--11 with BiMamba token mixers.
The tradeoff comparison in Table~\ref{tab:tradeoff} is organized under matched token-retention ratios.
Dense DeiT serves as the dense reference, A-ViT serves as the sparse-attention baseline, and the proposed model replaces only the selected late layers and is trained with the same sparsity-guided distillation procedure as in Sec.~4.2.
This setup ensures that the comparison reflects how the same sparsity budget is used, rather than differences in sparsity level.

\subsection{Throughput and Speedup Measurement}
\label{sec:app_setup_throughput}

Because A-ViT induces layer-dependent token counts, per-layer throughput is not directly comparable across layers.
In addition, the official A-ViT implementation does not provide token-routing code that realizes actual end-to-end acceleration.
We therefore report aggregated token-mixing throughput and an estimated model-level speedup.

\paragraph{Measured token throughput.}
The measured throughput in Table~\ref{tab:tradeoff} is reported in tokens/ms.
For each token-mixing layer $l$, let $T_l$ denote the number of processed tokens and $t_l$ the measured token-mixer runtime.
The reported throughput is computed as
\begin{equation}
\mathrm{Throughput} = \frac{\sum_l T_l}{\sum_l t_l},
\end{equation}
where the summation is taken over the token-mixing layers under evaluation.

\paragraph{Estimated model speedup.}
Model-level speedup is estimated by substituting the measured token-mixer runtime into the corresponding replaced layers while keeping the remaining backbone unchanged.
This estimate is used in Table~\ref{tab:tradeoff} to compare the relative runtime effect of sparse attention and sparse sequential replacement under matched sparsity budgets.

\paragraph{Resolutions.}
The main tradeoff table reports token throughput at both $224^2$ and $384^2$ resolution.
The higher-resolution setting is included to illustrate how the efficiency advantage evolves with token count.

\section{Metrics and Measurement Details}
\label{sec:app_metrics}

This section formalizes the analysis metrics used in Secs.~4.1--4.3.

\subsection{Token Throughput and Estimated Speedup}
\label{sec:app_metrics_throughput}

Because A-ViT induces layer-dependent token counts, throughput is aggregated over the evaluated token-mixing layers rather than compared layer by layer.
For each token-mixing layer $l$, let $T_l$ denote the number of processed tokens and $t_l$ the measured runtime of the token-mixing operator.
The reported token throughput is defined as
\begin{equation}
\mathrm{TP}=\frac{\sum_{l\in\mathcal{L}} T_l}{\sum_{l\in\mathcal{L}} t_l},
\label{eq:token_throughput}
\end{equation}
where $\mathcal{L}$ is the set of token-mixing layers under evaluation.
The unit of $\mathrm{TP}$ is tokens/ms.

Since the official A-ViT implementation does not provide token-routing code that realizes actual end-to-end acceleration, we additionally report an estimated model speedup.
Let $t_{\mathrm{mix}}=\sum_{l\in\mathcal{L}} t_l$ denote the aggregated token-mixing runtime and let $t_{\mathrm{fix}}$ denote the runtime of the unchanged backbone components.
The estimated model runtime is
\begin{equation}
t_{\mathrm{model}} = t_{\mathrm{fix}} + t_{\mathrm{mix}},
\label{eq:model_runtime}
\end{equation}
and the corresponding speedup relative to the attention-based baseline is
\begin{equation}
\mathrm{Speedup}=
\frac{t_{\mathrm{model}}^{\mathrm{attn}}}{t_{\mathrm{model}}^{\mathrm{method}}}.
\label{eq:model_speedup}
\end{equation}

\subsection{Gradient-Based Interaction Maps}
\label{sec:app_metrics_gradmap}

For attention-based models, token interactions are visualized directly by the attention scores.
For sequential replacements, which do not expose an explicit interaction matrix, we use gradient-based interaction maps.

Let $\mathbf{x}^{(l)}=[\mathbf{x}^{(l)}_1,\dots,\mathbf{x}^{(l)}_T]\in\mathbb{R}^{T\times D}$ denote the input tokens to layer $l$, and let $\mathbf{u}^{(l,h)}_i\in\mathbb{R}^{D_h}$ denote the output feature of token $i$ from head $h$ before the final post-projection.
The interaction score from input token $j$ to output token $i$ is defined as
\begin{equation}
M^{(l,h)}_{ij}
=
\sum_{d=1}^{D}
\left|
\frac{\partial \left(\sum_{k=1}^{D_h} u^{(l,h)}_{i,k}\right)}
{\partial x^{(l)}_{j,d}}
\right|.
\label{eq:grad_map}
\end{equation}
That is, we first sum the feature dimensions of the target output token, backpropagate to the layer input, and then aggregate the absolute gradient over the feature dimension of each input token.
In the reported visualizations, the interaction maps are averaged over the input batch.

\subsection{Token Importance and AUPRC}
\label{sec:app_metrics_importance}

To evaluate whether the tokens retained by A-ViT are informative, we compare them with token importance scores derived from the same dependency signal.
For a fixed layer $l$, the importance of input token $j$ is defined by aggregating its influence on all output tokens:
\begin{equation}
s^{(l)}_j = \sum_{i=1}^{T} M^{(l)}_{ij},
\label{eq:token_importance}
\end{equation}
where $M^{(l)}$ denotes the interaction map used for analysis.

Let $\mathbf{m}^{(l)}\in\{0,1\}^T$ denote the binary retention mask produced by A-ViT at layer $l$, where $m^{(l)}_j=1$ indicates that token $j$ is retained.
We quantify the alignment between retained tokens and token importance by treating $\mathbf{m}^{(l)}$ as binary labels and $\mathbf{s}^{(l)}=[s^{(l)}_1,\dots,s^{(l)}_T]$ as prediction scores, and then computing the area under the precision--recall curve:
\begin{equation}
\mathrm{AUPRC}^{(l)} = \mathrm{AUPRC}\!\left(\mathbf{m}^{(l)}, \mathbf{s}^{(l)}\right).
\label{eq:auprc}
\end{equation}
A higher $\mathrm{AUPRC}^{(l)}$ indicates stronger agreement between the tokens retained by A-ViT and the tokens ranked as important by the dependency-based score.

\end{document}